# Consciousness as a logically consistent and prognostic model of reality[1]

Evgenii Vityaev [1,2,*]

[1] Sobolev institute of mathematics, Koptuga 4, Novosibirsk, Russia; vityaev@math.nsc.ru
[2] Novosibirsk State University, Pirogova 2, Novosibirsk, Russia; vityaev@math.nsc.ru
[*] Correspondence: vityaev@math.nsc.ru; Tel.: +7-913-945-4867

**Abstract:** The work demonstrates that brain might reflect the outer world causal relationships in the form of a logically consistent and prognostic model of reality, which shows up as consciousness. The paper analyses and solves the problem of statistical ambiguity and provides a formal model of causal relationships as probabilistic maximally specific rules. We suppose that brain makes all possible inferences from causal relationships. We prove that the suggested formal model has a property of an unambiguous inference: from consistent premises we infer a consistent conclusion. It enables a set of all inferences to form a consistent model of the perceived world. Causal relationships may create fixed points of cyclic inter-predictable properties. We consider the "natural" classification introduced by John St. Mill and demonstrate that a variety of fixed points of the objects' attributes forms a "natural" classification of the outer world. Then we consider notions of "natural" categories and causal models of categories, introduced by Eleanor Rosch and Bob Rehder and demonstrate that fixed points of causal relationships between objects attributes, which we perceive, formalize these notions. If the "natural" classification describes the objects of the external world, and "natural" concepts the perception of these objects, then the theory of integrated information, introduced by G. Tononi, describes the information processes of the brain for "natural" concepts formation that reflects the "natural" classification. We argue that integrated information provides high accuracy of the objects identification. A computer-based experiment is provided that illustrates fixed points formation for coded digits.

## 1. Introduction

The work demonstrates that the human brain may reflect the outer world causality in the form of a logically consistent and prognostic model of reality that shows up as consciousness.

The work analyses and solves such problem of causal reflection of the outer world as a statistical ambiguity (Section 2.3). The problem is solved in such a way that it is possible to obtain a formal model of causal relationships, which provides a consistent and prognostic model of the outer world. To discover these causal relationships by the brain, a formal model of neuron that is in line with Hebb rule (Hebb, 1949), is suggested. We suppose that brain makes all possible inferences/predictions from those causal relationships. We prove (see Section 2.5) that the suggested formal model of causal relationships has a property of an unambiguous inference/predictions, namely, consistent implications are drawn out from consistent premises. It enables a set of all inferences/predictions, which brain makes from causal relationships, to form a consistent and predictive model of the perceived world. What is particularly important is that causal relationships may create fixed points of cyclic inter-predictable properties that create a certain "resonance" of inter-predictions. In terms of interconnections between neurons, these are cellular assemblies of neurons that mutually excite each other and form the systems of highly integrated information. In the formal model these are logically consistent fixed points of causal relationships. We argue (Section 2.1) that if attributes of the outer world objects are taken regardless of how persons perceive them, a complex of fixed points of the objects' attributes forms a "natural" classification of the outer world. If the fixed points of causal relationships of the outer world objects, which persons perceive, are taken, they form "natural" concepts described in cognitive sciences (Section 2.2).

[1] The work is supported by the Russian Science Foundation grant №17-11-01176 in part concerning the mathematical results in 2.4-2.6 and by Russian Foundation for Basic Research # 19-01-00331-a in other parts.

If the "natural" classification describes objects of the external world, and "natural" concepts are the perception of these objects, then the theory of integrated information (Tononi, 2004, 2016, Ozumi, Albantakis and Tononi, 2014) describes the information processes of the brain when these objects are perceived.

G. Tononi defines consciousness as a primary concept, which has the following phenomenological characteristics: composition, information, integration, exclusion (Ozumi, Albantakis and Tononi, 2014). For a more accurate determination of these properties G. Tononi introduces the concept of integrated information: "integrated information characterizing the reduction of uncertainty is the information, generated by the system that comes in a certain state after the causal interaction between its parts, which is superior information generated independently by its parts themselves" (Tononi, 2004).

The process of reflection of causal relationships of the outer world (Fig. 1) shall be further considered. It includes:
1. The objects of the outer world (cars, boats, berths) which relate to certain "natural" classes;
2. The process of brain reflection of objects by causal relations marked by blue lines;
3. Formation of the systems of interconnected causal relationships, indicated by green ovals.

In G. Tononi's theory only the third point of reflection is considered. The totality of the excited groups of neurons form a maximally integrated conceptual structure that defined by G. Tononi as qualia. Integrated information is also considered as a system of cyclic causality. Using integrated information the brain is adjusted to perceiving "natural" objects of the outer world.

In terms of integrated information, phenomenological properties are formulated as follows. In brackets an interpretation of these properties from the point of view of "natural" classification is given.
1. Composition – elementary mechanisms (causal relationships) can be combined into the higher-order ones ("natural" classes in the form of causal loops produce a hierarchy of "natural" classes);
2. Information – only mechanisms that specify "differences that make a difference" within a system shall be taken into account (only a system of "resonating" causal relationships, forming a class and "differences that make a difference" is important. See illustration in the computer experiment below);
3. Integration – only information irreducible to non-interdependent components shall be taken into account (only system of "resonating" causal relations, indicating an excess of information and perception of highly correlated structures of "natural" object is accounted for);
4. Exclusion – only maximum of integrated information counts (only values of attributes that are "resonating" at the fix-point and, thus, mostly interrelated by causal relationships, form a "natural" class or "natural" concept).

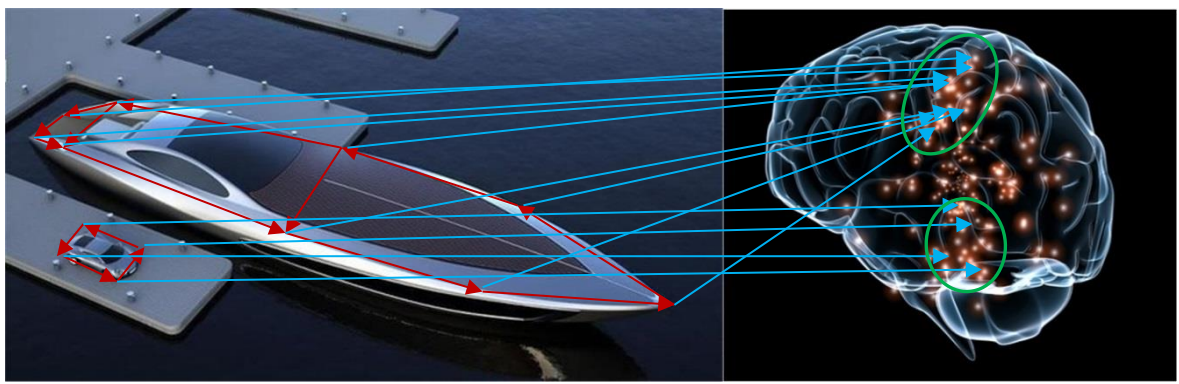

**Figure 1**. Brain reflection of causal relationships between objects attributes

These properties are defined as the intrinsic properties of the system. We consider these properties as the ability of the system to reflect the complexes of external objects' causal relations, and consciousness as the ability of a complex hierarchical reflection of a "natural" classification of the outer world.

Theoretical results on consistency of inference and consistency of fixed points of our formal model are supposing that a probability measure of events is known. However, if we discover causal relationships on the training set, and intend to predict properties of a new object out of the training set and belonging to a wider general population, or to recognize a new object as a member of some "natural" concept, there might be inconsistencies. Here, a certain criterion of maximum consistency is employed (see Section 2.8), which is based upon information measure, close in meaning to an entropic measure of integrated information (Tononi, 2004). The process of recognizing "natural" classes is described in Section 2.9.

In section 3 a computer modeling of "fixed points" discovering for the coded digits is provided.

Further we shell describe the idea of the work in more detail before the complicated mathematical part.

Causality is a result of physical determinism: "for every isolated physical system some fixed state of a system determines all the subsequent states" (Carnap, 1966). But, an automobile accident shall be taken as an example (Carnap, 1966). What was the reason for it? It might be a road surface condition or humidity, position of sun with respect to drivers' looks, reckless driving, psychological state of driver, functionality of brakes, etc. It is clear that there is no any certain cause in this case.

In the philosophy of science causality is reduced to forecasting and explaining. "Causal relation means predictability … in that if the entire previous situation is known, an event may be predicted …, if all the facts and laws of nature, related to the event, are given" (Carnap, 1966). It is clear that nobody knows all the facts, which number in case of an accident is potentially infinite, and all the laws of nature. In case of a human being and animals, the laws are obtained by training (inductive reasoning). Therefore, causality is reduced to predicting pursuant to inductive-statistical (I-S) reasoning, which involves logical inference of predictions from facts and statistical laws with some probabilistic assessment.

When discovering causal relationships and laws on real data or by training, a problem of statistical ambiguity appears – contradictions (contradictory predictions) may be inferred from these causal relationships. See example in the section 2.3. To avoid inconsistences, Hempel (1965, 1968) introduced a requirement of maximal specificity (see Sections 2.3, 2.4), which implies that a statistical law should incorporate maximum of information, related to the predictable property.

Section 2.5 presents a solution to the problem of statistical ambiguity. Following Hempel, a definition of maximally specific rules is given (Section 2.4, 2.5), for which it is proved (Section 2.6) that (I-S) inference that uses only maximum specific rules, does not result in inconsistencies (see also Vityaev, 2006). A special semantic probabilistic inference is developed that discover all maximum specific rules, which might be considered as the most precise causal relationships (that have maximum conditional probability and use maximum available information). Work (Vityaev, 2013) shows that the semantic probabilistic inference might be considered as a formal model of neuron that satisfy the Hebb rule, in which the semantic probabilistic reasoning discover all most precise conditional relationships. A set of such neurons might create a consistent and prognostic model of the outer world. Causal relationships may form fixed points of cyclic inter-predictable properties that produce a certain "resonance" of mutual predictions. Cycles of inferences about causal relations, are mathematically described by the "fixed points". These points are characterized by the property, that using inferences with respect to considered properties they don't predict some new property.

Neuron transmits its excitation to the other neurons through multiple both excitatory and inhibitory synapses. Inhibitory synapses may slow down other neurons and stop their activity. It is important for "inhibiting" alternate perception images, attributes, properties, etc. Within our formal model it is accomplished by discovering "inhibitory" conditional relationships that predict absence of an attribute/property of the object (the perceived object shall not have the respective attribute/property) as compared to the other objects, where this characteristic is found. A formal model specifies it by predicates' negations for corresponding attribute/property. Absence of inconsistencies at the fixed point means that there are no two causal relationships simultaneously predicting both availability of some attribute/property with an object, and its absence.

The structure of the outer world objects was analyzed in the form of "natural" classifications (Section 2.1). It was noted that "natural" classes of animals or plants have a potentially infinite number of different properties (Mill, 1983). Naturalists, who were building "natural" classifications, also noted that construction of a "natural" classification was just an indication: from an infinite number of attributes you need to pass to the limited number of them, which would replace all other attributes (Smirnof, 1938). This means that in "natural" classes these attributes are strongly correlated, for example, if there are 128 classes, and their attributes are binary, then the independent "indicator" attributes among them will be about 7 attributes as $2^7 = 128$, and others can be predicted based on these 7 attributes. The set of mutually predicted properties, obtained at the fixed point, produces some "natural" class.

High correlation of attributes for "natural" classes was also confirmed in cognitive studies (see. Section 3). Eleanor Rosch formulated the principles of categorization, one of which is as follows: "the perceived world is not an unstructured set of properties found with equal probability, on the contrary, the objects of perceived world have highly correlated structure" (Rosch, 1973, 1978, Rosch at al. 1976). Therefore, directly perceived objects (so called basic objects) are observed and functional ligaments, rich with information. Later, Bob Rehder suggested a theory of causal models, in which the relation of an object to a category is based not on a set of attributes but on the proximity of generating causal mechanisms: "the object is classified as a member of a certain category to the extent that its properties are likely to have been generated by this category of causal laws" (Rehder, 2003). Thus, the structure of causal relationships between the attributes of objects is taken as a basis of categorization. Therefore, brain perceives a "natural" object not as a set of attributes, but as a "resonant" system of causal relationships, closing upon itself through simultaneous inference of the total aggregate of the "natural" concept features. At the same time, "resonance" occurs, if and only if these causal relationships reflect some integrity of some "natural" class, in which a potentially infinite number of attributes mutually presuppose each other. To formalize causal models, Bob Rehder proposed to use causal graphical models (CGMs). However, these models are based on «deployment» of Bayesian networks, which do not allow cycles and cannot formalize cyclic causal relationships.

It should be specially noted that the "resonance" of mutual predictions of the properties of objects is carried out continuously in time and therefore the predicted properties are properties that have just been perceived as stimuli from the object. Therefore, it is always a prediction in time. The absence of contradictions in the predictions is also the absence of contradictions between the predicted stimulus and the really received stimulus.

Here we suggest a new mathematical apparatus for formalizing cyclic causal relationships.

## 2. Materials and Methods

### 2.1. "Natural" classification

The first fairly detailed philosophical analysis of "natural" classification was carried out by John St. Mill (1983). This analysis describes all the basic properties of "natural" classifications, which were later confirmed by naturalists who were building "natural" classifications and also by those researchers in the field of cognitive sciences, who were investigating "natural" concepts.

According to John St. Mill (1983) "artificial" classifications differ from the "natural" ones in that they may be based on any one or more attributes, so that different classes differ only in inclusion of objects, having different values of these attributes. But if the classes of "animals" or "plants" are considered, they differ by such a large (potentially infinite) number of properties that they can't be enumerated. Furthermore, all these properties are based on statements, confirming this distinction.

At the same time, among the properties of some "natural" classes, there are both observable and non-observable attributes. To take into account hidden attributes, their causal exhibitions in the observed attributes should be found. "For example, the natural classification of animals should be based mainly on their internal structure; however, it would be strange, as noted by A. Comte, if we were able to determine the family and specie of one or another animal, only by killing it" (Mill, 1983).

James. St. Mill gives the following definition of "natural" classification: it is such a classification, which combines the objects into the groups, about which we can make the greatest number of common propositions, and it is based on such properties, which are causes of many others … He also defines an "image" of a class, which is the forerunner of "natural" concepts: "our concept of class, the way in which this class is represented in our mind, is the concept of some sample, having all attributes of this class in the highest … degree".

John St. Mill's arguments were confirmed by naturalists. For example, Rutkowski (1884) writes about inexhaustible number of general properties of the "natural" classes: "The more essential attributes are similar in comparable objects, the more likely they are the similar in other attributes". Smirnov E.S. (1938) makes a similar statement: "The taxonomic problem is in indication: "from an infinite number of attributes we need to pass to a limited number of them, which would replace all other attributes." As a result, there were formulated a problem of specifying "natural" classifications, which is still being discussed in the literature (Wilkins and Ebach, 2013).

These researches are uncover a high degree of over determination of information for "natural" classes. If for specification of 128 classes with binary characteristics only 7 binary attributes are required, since $2^7 = 128$, then for hundreds or even thousands of attributes as it is the case of "natural" classes, they are much more highly overdetermined. Since over determination of features is high, it is possible to find, for example, 20 attributes for our classes, which are identical for the objects of the same class and entirely different in other classes. As each class is described by the fixed set of values of any 7 from 20 binary attributes, then remaining 13 attributes are uniquely determined. It implies that there are, at least, $13 * C_{20}^7 = 1007760$ dependencies between attributes. This property of "natural" classification was called as "taxonomic saturation" (Kogara, 1982). For such systems of attributes there is no a problem of a "curse of dimensionality", when as dimensionality of the attribute space increases, accuracy of Machine Learning algorithms decreases. Conversely it was shown in (Nedelko, 2015) that accuracy of the classification algorithm increases, if it properly incorporates redundancy of information. The classification and recognition algorithm in Sections 2.8, 2.9 incorporate this redundant information. As it is shown in the works on "natural" concepts in cognitive studies (see next section), human perception also incorporates this redundancy of information that allows identifying objects with practically with very high accuracy.

*2.2. Principles of categorization in cognitive sciences*

In the works of Eleanor Rosch (1973, 1978, Rosch at al. 1976) the principles of categorization of "natural" categories, confirming statements of John St. Mill and naturalists, are formulated on the basis of the experiments:

"Cognitive Economy. The first principle contains the almost common-sense notion that, as an organism, what one wishes to gain from one's categories is a great deal of information about the environment while conserving finite resources as much as possible. To categorize a stimulus means to consider it, for purposes of that categorization, not only equivalent to other stimuli in the same category but also different from stimuli not in that category. On the one hand, it would appear to the organism's advantage to have as many properties as possible predictable from knowing any one property, a principle that would lead to formation of large numbers of categories with as fine discriminations between categories as possible».

«Perceived World Structure. The second principle of categorization asserts that … perceived world – is not an unstructured total set of equiprobable co-occurring attributes. Rather, the material objects of the world are perceived to possess … high correlational structure. … In short, combinations of what we perceive as the attributes of real objects do not occur uniformly. Some pairs, triples, etc., are quite probable, appearing in combination sometimes with one, sometimes another attribute; others are rare; others logically cannot or empirically do not occur».

It is understood that the first principle is not possible without the second one – the cognitive economy is not possible without a structured world. «Natural» objects (basic objects) are rich with information ligaments of observed and functional properties, which form a natural discontinuity, creating categorization. These ligaments form a "prototypes" of the objects of different classes ("images" of John St. Mill): «Categories can be viewed in terms of their clear cases if the perceiver places emphasis on the correlational structure of perceived attributes … By prototypes of categories we have generally meant the clearest cases of category membership» (Rosch, 1973, 1978). «Rosch and Mervis have shown that the more prototypical of a category a member is rated, the more attributes it has in common with other members of the category and the fewer attributes in common with members of the contrasting categories» (Rosch & Mervis, at al., 1976).

For the future the theory of "natural" concepts was suggested by Eleanor Rosch, called the Prototypical Theory of Concepts (Prototype Theory). Its main features are described as follows: «The prototype view (or probabilistic view) keeps the attractive assumption that there is some underlying common set of features for category members but relaxes the requirement that every member have all the features. Instead, it assumes there is a probabilistic matching process: Members of the category have more features, perhaps weighted by importance, in common with the prototype of this category than with prototypes of other categories» (Ross, et al., 2008).

In further studies it was found that the models based on attributes, similarities and prototypes are not sufficient to describe classes. It is therefore necessary to take into account the theoretical, causal and ontological knowledge, relating to the objects of classes. For example, people do not only know that birds have wings and can fly, and build their nests in trees, but also facts, that the birds build their nests in trees, because they can fly, and fly because they have wings.

Many researchers believe that the most important theoretical knowledge is the knowledge of causal dependencies because of its functionality. It allows the organism to interfere in external events and to gain control over the environment. Studies have shown that people's knowledge of categories isn't limited by the list of properties, and includes a rich set of causal relationships between these properties, which can be ranked. The importance of these properties also depends on the category causal relationships. It was shown in some experiments (Ahn, at al., 2000, Sloman, Love and Ahn, 1998), that property is more important in classification, if it is more involved in causal network of relationships between the attributes. In the other experiments it was shown, that the property is more important, if it has more reasons (Rehder & Hastie, 2001).

Considering causal dependencies, Bob Rehder proposed the theory of causal models, according to which: «people's intuitive theories about categories of objects consist of a model of the category, in which both the category's features and the causal mechanisms among those features are explicitly represented. In other words, theories might make certain combinations of features either sensible and coherent … in light of the relations linking them, and the degree of coherence of a set of features might be an important factor determining membership in a category» (Rehder, 2003).

In the theory of causal models, the relationship of the object to a category is based not on a set of attributes or similarity based on attributes, but on the basis of similarity of generating causal mechanisms: «Specifically, a to-be-classified object is considered a category member to the extent that its features were likely to have been generated by the category's causal laws, such that combinations of features that are likely to be produced by a category's causal mechanisms are viewed as good category members and those unlikely to be produced by those mechanisms are viewed as poor category members. As a result, causal-model theory will assign a low degree of category membership to objects that have many broken feature correlations (i.e., cases where causes are present but effects absent or vice versa). Objects that have many preserved correlations (i.e., causes and effects both present or both absent) will receive a higher degree of category membership because it is just such objects that are likely to be generated by causal laws» (Rehder, 2003).

To represent causal knowledge, some researchers have used Bayesian networks and causal models (Cheng, 1997, Gopnik, at al., 2004, Griffiths &Tenenbaum, 2009). However, these models cannot simulate cyclic causality because Bayesian networks do not support cycles. In his work Bob Rehder (Rehder & Martin, 2011) proposed a model of causal cycles, based on a "disclosure" of causal graphical models. The "disclosure" is fulfilled by creating a Bayesian network which unfolds over time. But this work did not formalize the cycles of causality.

Our model is directly based on cyclic causal relationships, represented by fundamentally new mathematical models – fixed points of predictions on causations. To formalize such fixed points, a probabilistic generalization of formal concepts was defined (Vityaev, et al., 2012, 2015). Formal concepts emerging in the Formal Concept Analysis (FCA) may be specified as fixed points of deterministic implications (with no exceptions) (Ganter and Wille, 1999, Ganter, 2003). We generalize formal concepts for probabilistic case through introducing probabilistic implications and defining fixed points for probabilistic implications (Vityaev, et al., 2012, 2015). Generalization is made so, that in certain conditions probabilistic formal concepts and formal concepts coincide. A computer experiment is performed (Vityaev, et al., 2012, 2015), which demonstrates that probabilistic formal concepts might "restore" formal concepts after superimposition of noise.

The formalization given in the section 2.6 is more general, then in works (Vityaev, et al., 2012, 2015). When there is a fixed set of objects, and there is no general population, for which we need to recognize a new object from general population as belonging to one of available formal concepts, formalization in section 2.6 provides probabilistic formal concepts. But when set of objects is a sample from the general population, and it is necessary to recognize new objects to the one of the probabilistic formal concepts, this formalization provides "natural" classification of the objects of general population.

*2.3. Statistical ambiguity problem*

A problem of inconsistent predictions obtained from inductively deduced knowledge is called a problem of statistical ambiguity. The classical example is the following. Suppose that we have the following statements.

L1 Almost all cases of streptococcus infection clear up quickly after the administration of penicillin.

L2 Almost no cases of penicillin resistant streptococcus infection clear up quickly after the administration of penicillin.

C1 Jane Jones had streptococcus infection.

C2 Jane Jones received treatment with penicillin.

C3 Jane Jones had a penicillin resistant streptococcus infection.

On the base of L1 and C1∧C2 one can explain why Jane Jones recovered quickly (E). The second argument with L2 and C2∧C3 explains why Jane Jones did not (¬E). The set of statements {C1, C2, C3} is consistent. However, the conclusions contradict each other, making these arguments rival ones. Hempel hoped to solve this problem by forcing all statistical laws in an argument to be maximally specific – they should contain all relevant information with respect to the domain in question. In our example, then, the statement C3 invalidates the first argument L1, since this argument is not maximally specific with respect to all information about Jane Jones. So, we can only explain ¬E, but not E. Let us consider the problem of statistical ambiguity the notion of maximal specificity in more detail.

There are two types of predictions (explanations): deductive-nomological (D-N) and inductive-statistical (I-S). The employed laws in D-N predictions are supposed to be true, whereas in I-S predictions they are supposed to be statistical.

***Deductive-nomological model*** may be presented in the form of a scheme

$$\frac{\begin{array}{c}L_1,\ldots,L_m\\ C_1,\ldots,C_n\end{array}}{G}$$

where: $L_1,\ldots,L_m$ - set of laws; $C_1,\ldots,C_n$ - set of facts; G – predicted statement; $L_1,\ldots,L_m, C_1,\ldots,C_n \vdash G$; set $L_1,\ldots,L_m, C_1,\ldots,C_n$ is consistent; $L_1,\ldots,L_m \nvdash G$, $C_1,\ldots,C_n \nvdash G$; laws $L_1,\ldots,L_m$ contain only universal quantifiers; set of facts $C_1,\ldots,C_n$ – quantifier-free formulas.

***Inductive-statistical model*** is identical to deductive-nomological with the difference that laws are statistical and shall meet the Requirement of Maximal Specificity (RMS), to avoid in-consistencies.

Requirement of Maximal Specificity is defined by Carl Hempel (1965, 1968) as follows:

$$\frac{\begin{array}{c}F \Rightarrow G,\ p(G;F) = r\\ F(a)\end{array}}{G(a)}[r].$$

Rule $F \Rightarrow G$ is the maximum specific with the state of knowledge K, if for each class H, for which both of the below statements belong to K

$\forall x(H(x) \Rightarrow F(x))$, $H(a)$

there is a statistical law $p(G;H) = r'$ in K such that $r = r'$. RMS requirement implies that if both F and H contain object *a*, and H is a subset of F, then H has more specific information about object *a*, than F and, therefore, law $p(G;H)$ shall prevail over law $p(G;F)$. However, for maximum specific rules law $p(G;H)$ should have the same probability as law $p(G;F)$ and thus H not adds any additional information.

The Requirement of Maximal Specificity had not been investigated by Hempel and its successors formally, and it had not been proved that it can avoid inconsistencies. The next section contains a formal definition of the maximal specificity, for which we prove that I-S inference that use only maximal specific rules, is consistent.

*2.4. Requirement of Maximal Specificity*

Let us introduce a language of the first order L of signature $\Im = \langle P_1,\ldots,P_k \rangle$, which contains only unitary predicates symbols for the objects properties and stimulus description. Let $U(\Im)$ shall denote a set of all atomic formulas. Atomic formulas or their negations shall be called *liters*, and a multitude of all liters shall be denoted as Lit. Closure of all liters with respect to logical operations $\&, \vee, \neg$ shall be called a set of sentences $\Re(\Im)$.

Also we need an empirical system $M = \langle A, W \rangle$ of the signature $\Im$ for representing the set of objects A and set of predicates $W = \langle P_1,\ldots,P_k \rangle$ defined on A. Set of all M-true sentences from $\Re(\Im)$ shall be called a theory Th(M) of M. It shall be further supposed that theory Th(M) – is a set of universally quantified formulas. It is known that any set of universal formulas is logically equivalent to the set of rules as follows

$$C = (A_1 \& \ldots \& A_k \Rightarrow A_0),\ k \geq 0,\ A_0 \notin \{A_1,\ldots,A_k\}, \tag{1}$$

where $A_0, A_1,\ldots,A_k$ – liters. Formulas of type $A_0$, $k = 0$ are considered as rules ($T \Rightarrow A_0$), where T – truth. Hence, without loss of generality, it can be assumed that theory Th(M) is a set of the (1) type rules.

A rule might be true on empirical system M only because the premise of the rule is always false. Furthermore, a rule may also be true, since its certain "sub-rule" that involves a part of premise, is true on empirical system. These observations are summarized in the following theorem.

**Теорема 1** 8. (Vityaev, 2006). Rule $C = (A_1 \& ... \& A_k \Rightarrow A_0)$ logically follows from rules:

1. $(A_{i_1} \& ... \& A_{i_h} \Rightarrow \neg A_{i_0}) \vdash C$, $\{A_{i_1},...,A_{i_h}, A_{i_0}\} \subset \{A_1,...,A_k\}$, $0 \le h < k$;

2. $(A_{i_1} \& ... \& A_{i_h} \Rightarrow A_0) \vdash C$, $\{A_{i_1},...,A_{i_h}\} \subset \{A_1,...,A_k\}$, $0 \le h < k$,

where $\vdash$ is prof.

**Definition 1**. *Sub-rule* of rule C shall be any rule of 1 or 2 type, specified in theorem 1.

**Corollary 1.** If sub-rule of rule C is true on M, then rule C is true on M.

**Definition 2**. Any rule C, true on M, each sub-rule of which is already not true on M, shall be called a *law* of empirical system M. Rule $(\Rightarrow A_0)$ is true on M, if $M \vDash A_0$. Rule $(\Rightarrow A_0)$ true on M, is a law on M.

Let Law – be the set of all laws on M. Then, theory Th(M) logically follows from Law.

**Theorem 2** (Vityaev, 2006). (See prof in Appendix A). Law $\vdash$ Th(M) and for each rule $C \in$ Th(M) its sub-rule exists, which is a law on M.

Statistical laws shall now be defined. For the sake of simplicity a probability shall be determined on empirical system $M = \langle A, W \rangle$ as on a general population, where A is a set of objects of a general population, and W is a set of unitary predicates, defined on A.

*Probability* $\mu : A \to [0,1]$ shall be defined on A (general case is considered in Halpern, 1990):

$$\sum_{a \in A} \mu(a) = 1, \mu(a) \ne 0, \ a \in A.$$

$$\mu(B) = \sum_{b \in B} \mu(b), \ B \subseteq A. \tag{2}$$

Probability $\mu^n$ on $(A)^n$, shall be naturally defined:

$\mu(B) = \sum_{b \in B} \mu(b)$ $\quad \mu^n(a_1,...,a_n) = \mu(a_1) \times ... \times \mu(a_n)$.

*Interpretation* of language L shall be determined as mapping $I : \mathfrak{I} \to W$, where predicate $P_j \in W$, $j = 1,...,k$ corresponds to each predicate symbol $P_j \in \mathfrak{I}$. Mapping $v : X \to A$ of a set of variables X of language L to the set of objects shall be called *attribution*. Composition of mappings $vI\varphi$, where $\varphi \in \mathfrak{R}(\mathfrak{I})$ specifies a formula obtained from φ by replacement of predicate symbols of by predicates W through interpretation I and replacement of variables from φ with objects from A by attributing v. Probability η of sentence $\varphi(a,...,b) \in \mathfrak{R}(\mathfrak{I})$ shall be defined as:

$$\eta(\varphi) = \mu^n(\{(a_1,...,a_n) \mid M \vDash vI\varphi, v(a) = a_1,...,v(b) = a_n\}). \tag{3}$$

**Definition 3.** Rule $C = (A_1 \& ... \& A_k \Rightarrow A_0)$ of (1) type, with $\eta(A_1 \& ... \& A_k) > 0$ and conditional probability $\eta(C) = \eta(A_0 / A_1 \& ... \& A_k) > 0$ strictly higher than conditional probabilities of all its sub-rules, shall be called a *probabilistic law* on M. Conditional probability of sub-rule $C = (\Rightarrow A_0)$ shall be defined as $\eta(C) = \eta(A_0 / T) = \eta(A_0)$. All rules of type $(\Rightarrow A_0)$, $\eta(A_0) > 0$ are probabilistic laws.

A set of all probabilistic laws shall be designated as LP.

**Definition 4.** Probabilistic law $C = (A_1 \& ... \& A_k \Rightarrow A_0)$, which is not a sub-rule of any other probabilistic law, shall be called the *strongest probabilistic law* (SPL-rule) on M. A set of all SPL-rules shall be designated as SPL.

It shall be proved that a concept of probabilistic law generalizes a concept of a law, which is true on M.

**Theorem 3.** (See prof in Appendix A). Law $\subset$ SPL $\subset$ LP.

**Definition 5** (Vityaev, 2006). *A semantic probabilistic inference* (SP-inference) of some SPL-rule, predicting liter $A_0$, shall be such sequence of probabilistic laws $C_1 \sqsubset C_2 \sqsubset ... \sqsubset C_n$, as:

1. $C_1 = (\Rightarrow A_0)$;

2. $C_1, C_2, ..., C_n \in LP$, $C_i = (A_1^i \& ... \& A_{k_i}^i \Rightarrow A_0)$, $k_i \geq 0$;

3. $C_i$ is a sub-rule of $C_{i+1}$, i.e. $\{A_1^i \& ... \& A_{k_i}^i\} \subset \{A_1^{i+1} \& ... \& A_{k_{i+1}}^{i+1}\}$, $k_i < k_{i+1}$; (4)

4. $\eta(C_{i+1}) > \eta(C_i)$;

5. $C_n$ – SPL-rule.

A set of all SP-inferences, predicting liter $A_0$ shall be considered. This set may be presented as a *semantic probabilistic inference tree* for liter $A_0$.

**Lemma 2.** (See prof in Appendix A). Any probabilistic law $C = (A_1 \& ... \& A_k \Rightarrow A_0)$ belongs to some semantic probabilistic inference, which predicts liter $A_0$ and, hence, to the tree of semantic probabilistic inference of liter $A_0$.

**Definition 6.** *A maximum specific rule* $MS(A_0)$ on M for predicting liter $A_0$ shall be an SPL-rule that has a maximum value of conditional probability among all SPL-rules of a semantic probabilistic inference tree for liter $A_0$. If there are several rules with identical maximum value, all of them are maximum specific.

A set of all maximum specific rules shall be denoted as MSR.

**Proposition 1.** Law $\subset$ MSR $\subset$ SPL $\subset$ LP.

*2.5. Resolving the problem of statistical ambiguity*

A requirement of maximal specificity shall be defined in the general case. It shall be supposed that statement H in the Hempel's formulation of the requirement of maximal specificity is a sentence $H \in \Re(\Im)$ of the language L.

**Definition 7.** Rule $C = (F \Rightarrow G), F \in \Re(\Im)$, $G \in Lit$ meets the requirement of maximal specificity (RMS), if it follows from $H \in \Re(\Im)$ and $F(a) \& H(a)$ for some $a \in A$ that rule $C' = (F \& H \Rightarrow G)$ has the same probability $\eta(G / F \& H) = \eta(G / F) = r$.

In other words, RMS states that there is no sentence $H \in \Re(\Im)$, which would increase (or decrease, see the Lemma below) conditional probability $\eta(G / F) = r$ of the rule by adding it to the premise of the rule.

**Lemma 3.** (See prof in Appendix A). If statement $H \in \Re(\Im)$ decreases probability of rule i.e. $\eta(G / F \& H) < \eta(G / F)$ then the statement $\neg H$ increases it and $\eta(G / F \& \neg H) > \eta(G / F)$.

**Lemma 4.** (See prof in Appendix A). For each rule $C = (A_1 \& ... \& A_k \Rightarrow A_0)$ of form (1), there shall be found a probabilistic law $C' = (B_1 \& ... \& B_{k'} \Rightarrow A_0)$, $k' < k$, on M, for which $\mu(C') \geq \mu(C)$.

**Теорема 4** (Vityaev, 2006). (See prof in Appendix A). Any maximum specific rule $MS(G) = (F \Rightarrow G)$, $F \in \Re(\Im), G \in Lit$ meets the requirement of maximal specificity.

*2.6. Fixed points of predictions based on MSR rules. Solution of a statistical ambiguity problem.*

It shall be proved that any I-S inference is consistent for any set of rules $\{L_1,\ldots,L_m\} \subset \text{MSR}$. To do it, an inference by MSR rules and fixed points of inference as per MSR rules shall be considered.

**Definition 8.** An immediate successor operator Pr shall be defined by rules from $P \subset \text{MSR}$ with a set of liters L as follows:

$$\Pr_P(L) = L \cup \{A_0 \mid C = (A_1 \& \ldots \& A_k \Rightarrow A_0), \{A_1,\ldots,A_k\} \subset L, C \in P\}$$

**Definition 9.** A fixed point of operator Pr of immediate successor with respect to a set of liters L shall be a closure of this set of liters with respect to the immediate successor operator $\Pr_P^\infty(L)$, whence it follows that $\Pr_P(\Pr_P^\infty(L)) = L$.

**Definition 10.** A set of liters $L = \{L_1,\ldots,L_k\}$ is called a *compatible set*, if $\eta(L_1 \& \ldots \& L_k) > 0$.

**Definition 11.** A set of liters L is *consistent*, if it does not contain simultaneously atom $G$ and its negation $\neg G$.

**Proposition 2.** (See prof in Appendix A). If L is compatible, L is consistent.

It shall be shown that an immediate prediction retains the property of compatibility.

**Теорема 5** (Vityaev and Martinovich 2015). (See prof in Appendix A). If L is compatible, $\Pr_P(L)$ is also compatible, $P \subset \text{MSR}$.

To prove the theorem, the following lemma shall first be proved.

**Лемма 5.** (See prof in Appendix A). If for rules $A = (\bar{A} \Rightarrow G)$, $B = (\bar{B} \Rightarrow \neg G)$, $\bar{A} = A_1 \& \ldots \& A_k$, $\bar{B} = B_1 \& \ldots \& B_m$, $\eta(\bar{A} \& \neg \bar{B}) > 0$, $k \geq 0$, $m > 0$ inequality $\eta(G / \bar{A} \& \neg \bar{B}) > \eta(G / \bar{A})$ is valid, a rule exists that has a strictly higher conditional probability than rule A.

**Corollary 2.** If L is compatible, then $\Pr_P(L)$ is consistent for $P \subset \text{MSR}$.

**Corollary 3**. (Solution to a problem of statistical ambiguity). I-S inference is consistent for any set of laws $P = \{L_1,\ldots,L_m\} \subset \text{MSR}$ and set of facts $\{C_1,\ldots,C_n\}$.

**Corollary 4.** Fixed points $\Pr_P^\infty(L)$ for a compatible set of liters L are compatible and consistent.

A set of all fixed points $L = \Pr_{MSR}^\infty(N)$, obtained by all maximum specific rules on all compatible sets of liters N, shall be designated using Class(M).

*2.7. "Natural" classification as a fixed points of prediction by a maximum specific rules.*

Probabilistic formal concepts and "natural" classification shall be specified, using fixed points pursuant to maximum specific rules. If empirical system $M = \langle A, W \rangle$ is defined on some fixed set of objects A, the below defined probabilistic formal concepts and "natural" classifications for this set are in agreement. It can be shown (Vityaev et al., 2005, Vityaev & Martynovich, 2015), that this definition of "natural" classification satisfies all the requirements, which natural scientists imposed on "natural" classification.

**Definition 12.** Probabilistic formal concepts and "natrual" classification (Vityaev & Martynovich, 2015).

1. A set of all fixed points of Class(M) shall be called a set of all probabilistic formal concepts and "natural" classes of this empirical system M.
2. Each class/concept L specifies in empirical system M a set of objects, which belong to a class/concept $M(L) = \{b \in A \mid B_b \vDash L\}$, where $B_b = \langle b, W \rangle$ is a sub-model of model $M = \langle A, W \rangle$ generated by object $b \in A$.
3. A lawful model of class/concept $C = \langle L, Z_L \rangle$ shall be defined as a set of liters $L \in \text{Class}(M)$ of fixed point and a set of rules $Z_L \subset MSR$, applicable to liters from L.

4. A generating set of some class/concept $C = \langle L, Z_L \rangle$ shall be such a subset of liters $N \subset L$, that $L = \Pr_{MSR}^{\infty}(N)$.

5. A set S of atomic sentences $P_j(a)$, j=1,…,s, $s \leq k$ shall be called a system-forming, if for each class from Class(M) there is a generating set of liters, which are obtained from the system-forming set of atomic sentences by taking or not taking the negation.

6. A systematics shall be defined as a set $\Sigma = \langle S, Z_S, \{Z_{L_i}\}_{L_i \in Class(M)} \rangle$, where S is a system-forming set of atomic sentences, $Z_S$ – a law of systematization, which defines an order of taking negations for atomic sentences from S, $\{Z_{L_i}\}_{L_i \in Class(M)}$, a set of the rules for fixed points from Class(M).

*2.8. Method of "natural" data classification*

In the case of "natural" classification, an empirical system $M = \langle A, W \rangle$ as a general population, is unknown. Only a data sampled from a general population are known. Therefore, to develop a method for "natural" classification, the questions of making "natural" classification and fixed points by data samplings should be considered. By a sampling from a general population we shall mean a sub-model, $B_B = \langle B, W \rangle$, where B is a set of objects randomly selected from general population A. A frequency probability $\mu_B$ shall be specified within a sampling, by assuming $\mu_B(a) = 1/N, N = |B|$, according to which probability $\eta_B$ on sentences from $\Re(\Im)$ is determined. On sampling $B_B = \langle B, W \rangle$, as on a sub-model, a theory Th(B), a set of laws Law(B), a set of probabilistic laws LP(B), a set of the strongest probabilistic laws SPL(B), and a set of maximum specific laws MSR(B) might be obtained.

**Proposition 3.** $Th(M) \subset Th(B)$.

Since each set of liters, $S_b = \{L \in Lit \mid B_b \vDash L\}$ is compatible, since it is obtained on real objects, which have a nonzero probability, hence, a set $Class(B) = \{L = \Pr_{MSR}^{\infty}(N) \mid N \subset S_b, b \in B\}$ shall be a set of all fixed points on sampling $B_B$. Classes Class(B) shall be the analogues of probabilistic formal concepts (Vityaev, et al., 2012, 2015). However, a potential identification and recognition of "natural" classes of a general population by "natural" classes discovered on a sampling, is of concern, rather than probabilistic formal concepts, defined on sampling $B_B \subset M$. In this case, probability $\mu$ of a general population is unknown, but frequency probability $\mu_B$ within a sampling is known.

Maximum specific laws MSR(B) within sampling $B_B$ may be not the same on a general population. They can be considered as approximations of laws MSR(M) in the following sense. Determining maximum specific laws using semantic probabilistic inference shall be recalled. In the process of inferring a premise of a rule is stepwise developed through strictly enhancing its conditional probability and involving as much relevant information as possible, on accordance with a requirement of maximal specificity, to ensure maximum probable and consistent prediction. Sampling $B_B$ might facilitate building a tree of semantic probabilistic inference through increasing the premise and applying some statistical criteria for checking the strict increase of the conditional probability $\mu$ on general population $M = \langle A, W \rangle$. For this purpose, an exact independence criterion of Fischer for contingency tables shall be used. Here, by sampling $B_B = \langle B, W \rangle$ a set $LP_\alpha(M)$ of probabilistic laws with some confidence level α might be discovered, where each probabilistic inequality shall be statistically validated with confidence level α. By the set $LP_\alpha(M)$, a set of the strongest probabilistic laws $SPL_\alpha(M)$ might be found, with confidence level α.

Inference by $SPL_\alpha(M)$ rules may be inconsistent, hence, to build fixed points by set $SPL_\alpha(M)$, it is necessary to use a weaker criterion of consistency of probabilistic laws in mutual predictions, which assumes occurrence of inconsistencies.

An operator of direct inference $Pr\Phi_{SPL_\alpha(M)}(L)$ shall be defined for this case. A set of laws, verified within a set of liters L shall be defined

$$Sat(L) = \{C \mid C \in SPL_\alpha(M), C = (A_1 \& ... \& A_k \Rightarrow A_0), \{A_1,...,A_k\} \subset L, A_0 \in L\},$$

and a set of laws, disprovable at a set of liters L

$$Fal(L) = \{C \mid C \in SPL_\alpha(M), C = (A_1 \& ... \& A_k \Rightarrow A_0), \{A_1,...,A_k\} \subset L, \neg A_0 \in L\}, (\neg\neg A_0 = A_0).$$

A criterion Kr of mutual consistency between laws from $SPL_\alpha$ shall be defined on a set of liters L as:

$$Kr_{SPL_\alpha(M)}(L) = \sum_{C \in Sat(L)} \nu(C) - \sum_{C \in Fal(L)} \nu(C), \text{ where } \nu(C) = -\log(1 - \eta_B(C)).$$

Function $-\log(1-\eta_B(C))$ incorporates not a probability itself, but its closeness to 1, since it characterizes a predictive power of regularity more accurately.

Criterion $Kr_{SPL_\alpha(M)}$ specifies not only information measure of consistency between regularities, but also an information measure of mutual integration between causal relationships within a set of liters L. Therefore, this measure is very close in meaning to entropy measure of integrated information stated in (Tononi, 2004).

Operator $Pr\Phi_{SPL_\alpha(M)}(L)$ functions as follows: it either adds one of liters Lit to a set L, which is predicted by regularities from $Sat(L)$, or deletes one of liters of a set L, predictable by disprovable laws from $Fal(L)$. Here, a compatibility of laws (their information measure) applicable to a set L shall strictly increase, i.e. at every step an inequality shall be satisfied

$$Kr_{SPL_\alpha}(Pr\Phi_{SPL_\alpha}(L)) > Kr_{SPL_\alpha}(L).$$

Here, that liter is added/deleted, which results in a maximum increase in the criterion (information measure). If adding/deleting a liter does not increase a criterion, a set L remains unchanged and is a fixed point. Changes in criterion, when adding/deleting an element, are, respectively:

$$\delta^+(L) = \max_{A_0 \in Pr_{SPL_\alpha}(L),\, A_0 \notin L} \{Kr_{SPL_\alpha(M)}(L \cup A_0) - Kr_{SPL_\alpha(M)}(L)\},$$

$$\delta^-(L) = \max_{A_0 \in Pr_{SPL_\alpha}(L),\, A_0 \in L} \{Kr_{SPL_\alpha(M)}(L \setminus A_0) - Kr_{SPL_\alpha(M)}(L)\}.$$

Operator $Pr\Phi_{SPL_\alpha(M)}(L)$ adds/deletes that element, which maximizes the value of a criterion. Added/deleted elements are defined as follows:

$$(A_0)^+ = \underset{A_0 \in Pr_{SPL_\alpha(M)}(L),\, A_0 \notin L}{\arg\max} (Kr_{SPL_\alpha(M)}(L \cup A_0)),$$

$$(A_0)^- = \underset{A_0 \in Pr_{SPL_\alpha(M)}(L),\, A_0 \in L}{\arg\max} (Kr_{SPL_\alpha(M)}(L \setminus A_0)).$$

In each case of employing, operator $Pr\Phi_{SPL_\alpha(M)}(L)$ adds/deletes that element, which increases a criterion to the maximum, i.e. adds element $(A_0)^+$, if $\delta^+(L) > \delta^-(L)$, $\delta^+(L) > 0$ and deletes element $(A_0)^-$, if $\delta^-(L) > \delta^+(L)$, $\delta^-(L) > 0$.

Thus, operator $Pr\Phi_{SPL_\alpha(M)}(L)$ is determined as follows:

$$Pr\Phi_{SPL_\alpha(M)}(L) = \begin{cases} L \cup (A_0)^+, \text{ if } \delta^+(L) > \delta^-(L), \delta^+(L) > 0 \\ L \setminus (A_0)^-, \text{ if } \delta^-(L) \geq \delta^+(L), \delta^-(L) > 0 \\ L, \text{ else} \end{cases}.$$

Fixed point $Pr\Phi_{SPL_\alpha(M)}(L) = L$ shall be obtained in the third case, when adding/deleting an element does not increase the criterion. A set of all such fixed points obtained on all compatible sets of liters L, shall be defined as a set of "natural" classes $Class_\alpha(M)$.

To obtain a regular model of class $L \in Class_\alpha(M)$ a set of regularities $Z_L$ interpredicting the class attributes should be specified. Regularities $Sat(L)$ represent such regularities. At the fixed point L disprovable predictions across regularities $Fal(L)$ are overlapped with the verified predictions across regularities $Sat(L)$. Since, theoretically, with the known measure $\mu$, there shouldn't be any inconsistencies, regularities $Fal(L)$ shall be considered obtained due to randomness of sampling $B_B$. Regularities from $SPL_\alpha(M)$ that were not included into any set $Sat(L)$ of any class $L \in Class_\alpha(M)$, are considered as obtained as a result of retraining by virtue of sampling $B_B$ randomness, therefore, they will be deleted from $SPL_\alpha(M)$.

**Definition 13**. A set $C = \langle L, Sat(L) \rangle$ shall be called a regular model of class $L \in Class_\alpha(M)$.

For "natural" classes, definitions of generating set, system forming set, and systematics from definition 12 remain unchanged.

The functions of information measure in terms of accuracy of identifying objects of the outer world shall be noted:
1. It enables to cope with retraining and exclude regularities, which are probably obtained due to randomness of sampling;
2. Extraction of the maximum value of information measure facilitates extracting all overdetermined information, inherent in "natural" classes;
3. Recognition of objects of the outer world using overdetermined information and makes this recognition maximally accurate and solves the problem of "dimensionality curse".

*2.9. Recognition of "natural" classes*

Regular models of classes allow recognizing them on control objects, chosen from a general population. For this purpose, regular models of classes $C = \langle L, Sat(L) \rangle$ in the form of regular matrices shall be defined. For each liter $A_0 \in L$ the power of its prediction shall be estimated pursuant to regularities from $Sat(L)$ and liters from L as

$$Kr_{Sat(L)}(A_0) = \sum_{C \in Sat(L),\, C=(A_1 \&\ldots\& A_k \Rightarrow A_0)} v(C).$$

**Definition 14**. Regular matrix of class $C = \langle L, Sat(L) \rangle$ shall be defined as a tuple $M_C = \langle L, \{Kr_{Sat(L)}(A_0)\}_{A_0 \in L} \rangle$.

**Proposition 4.** For each class $C = \langle L, Sat(L) \rangle$ an equation is true

$$Kr_{SPL_\alpha(M)}(L) = \sum_{A_0 \in L} Kr_{Sat(L)}(A_0).$$

Using regular matrices, new objects of a general population may be recognized and relegated to the "natural" classes. Pertinence of object $b \in B$ to class $C = \langle L, Sat(L) \rangle$ shall be assessed. To do that, it shall be estimated, to what degree object $b \in B$ is conform with regularities of class $Sat(L)$.

**Definition 15**. Pertinence of object $b \in B$ to class $C = \langle L, Sat(L) \rangle$ shall be defined as

$$Score(b/C) = \sum_{A_0 \in (S_b \cap L)} Kr_{Sat(L)}(A_0) - \sum_{A_0 \in L,\, \neg A_0 \in S_b} Kr_{Sat(L)}(A_0).$$

Recognition by regular matrices is made the same way as by weight matrices. To do that, for all objects of positive and negative sampling shall be calculated the $Score(b/C)$ for each class, and a threshold value shall be computed that provides the required values of the first and second type errors. Then new objects of a general set might be recognized and relegated to the class, if value $Score(b/C)$ of this object is higher than the threshold value.

## 3. Results

Let's illustrate the formation of fixed points for the "natural" classes/concepts by the computer experiment on coded digits. Digits shall be encoded, as it is shown in Fig. 2. Attributes of digits shall be enumerated, as indicated in Table 1. A training set, consisting of 360 shuffled digits (12 digits of Fig. 2, which are duplicated in 30 copies without specifying where any digit is) shall be formed. On this set 55089 maximum specific rules were found by semantic probabilistic inference. They are general statements about the objects, mentioned by John St. Mill and W. Whewell.

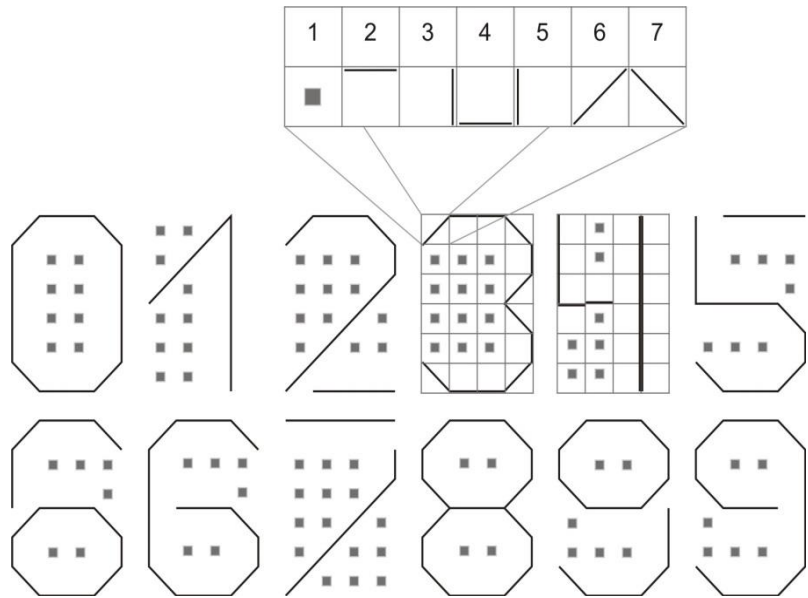

**Figure 2.** Digits coding.

Let $X(a)$ be a set of properties (stimulus) of an object *a*, defined by a set of predicates, and $(P_{i_1} \& ... \& P_{i_k} \Rightarrow P_{i_0}) \in MSR_\alpha(X)$ – the set of the maximum specific rules satisfied for properties X, $\{P_{i_1},...,P_{i_k}\} \subset X$ with confidential level α. Then we can define operator $Pr\Phi_{MSR_\alpha(X)}$ of direct inference. The fixed point is reached when $Pr\Phi_{MSR_\alpha(X)}^{n+1}(X(a)) = Pr\Phi_{MSR_\alpha(X)}^{n}(X(a))$, for some n, where $Pr\Phi_{MSR_\alpha(X)}^{n}$ – n multiple application of the operator. Since in each application of operator $Pr\Phi_{MSR_\alpha(X)}$ the value of criterion $Kr_{MSR_\alpha(X)}$ increases and at the fixed point reaches a local maximum, then the fixed point, when reflecting some "natural" object, has a maximum information measure $Kr_{MSR_\alpha(X)}$ and property of «exclusion» by G. Tononi.

**Table 1.** Encoding of digital's fields.

| 1 | 2 | 3 | 4 |
|---|---|---|---|
| 5 | 6 | 7 | 8 |
| 9 | 10 | 11 | 12 |
| 13 | 14 | 15 | 16 |
| 17 | 18 | 19 | 20 |
| 21 | 22 | 23 | 24 |
| Fields codes. | | | |

According to these regularities, exactly 12 fixed points are discovered, accurately corresponding to the digits. An example of a fixed point for digit 6 is shown in Figure. 3. It shall be considered, what this fixed point is.

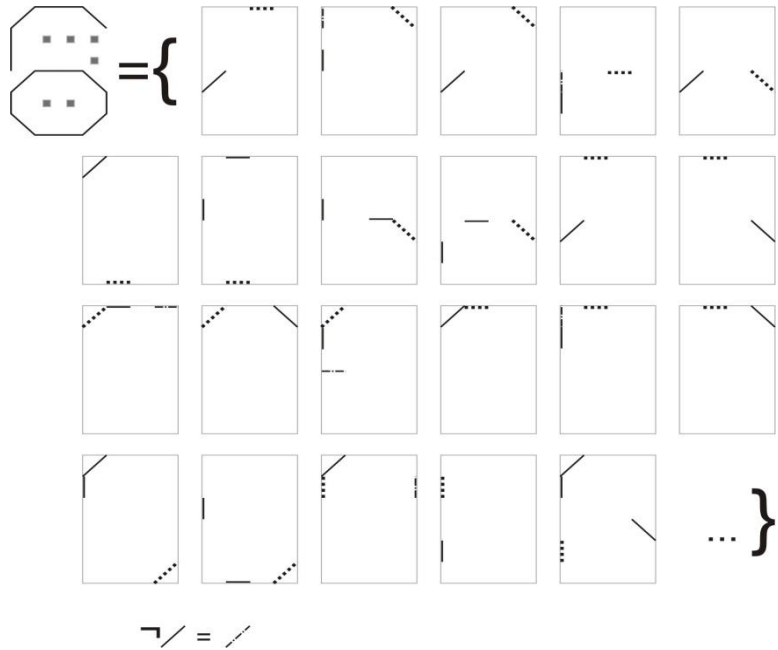

**Figure 3.** Fix-point for digit 6.

The first regularity of digit 6 in Figure 3, represented in the first box after the brace, means that if in the square 13 (Table. 1) there is an attribute 6 (it is denote as 13-6), then it must be an attribute 2 in square 3 (it is denote by (3-2) ). The predicted attribute is indicated by a dotted line. This regularity shall be written as (13-6 $\Rightarrow$ 3-2). It is easy to see that this regularity is really implemented in all the digits. The second regularity means, that from the attribute (9-5) and the denial of the value 5 of the first attribute ¬(1-5) (the first attribute must not be equal to 5) follows attribute (4-7). Denial is denoted by a dash line, as shown in the lower part of Figure 3. Thus, regularity (9-5&¬(1-5) $\Rightarrow$ 4-7) is obtained. The following 3 regularities in the first row of the digit 6, will be (13-6 $\Rightarrow$ 4-7), (17-5&¬(13-5) $\Rightarrow$ 4-7), (13-6 $\Rightarrow$ 16-7), respectively.

Figure 3 shows, that these regularities and the attributes of digit 6 form a fixed point. They mutually predict each other. It is worth mentioning, that regularities, used in the fixed point, are discovered on all digits, but fixed point allocates only one digit. This illustrates the phenomenological property of "information" which states that "differences that make a difference". Thus, the system of causal relationships perceives (is conscious of) whole object. Thus, digits are identified not by regularities themselves, but their system relationship. The fixed points forms a "prototype" by Eleanor Rosch, or "image" by John St. Mill. The program does not know in advance which possible combinations of attributes are correlated with each other.

**4. Discussion**

Theoretical results obtained in the paper suggest that it is possible to create a mathematically precise system of reflection of reality, based on the most specific rules and fixed points that use them. First of all, the question arises about the functioning of the neuron – is it really detects causal relations in accordance with the semantic probabilistic inference. Unfortunately, this is still impossible to verify, because connecting several contacts to one neuron is a deadly number for the neuron itself. However, it can be shown that the reflection of causal relationships is able to model a multitude of cognitive functions in accordance with existing physiological and psychological theories. The organization of purposeful behavior is modeled by causal relationships between actions and their results (Vityaev, 2015), which fully corresponds to the theory of functional systems (Anokhin, 1974). Fixed points adequately model the process of perception (Vityaev & Neupokoev,

2014). A set of causal relationships models expert knowledge (Vityaev, Perlovsky Kovalerchuk, Speransky, 2013). Therefore, the verification of this formal model for compliance with the actual processes of the brain seems to be an important task.

**Acknowledgments:** The authors thank the editors of the special issue "Integrated Information Theory" for publishing this paper in open access.

**Conflicts of Interest:** The author declares no conflict of interest.

**Appendix A. Mathematical proofs.**

*Proof of the theorem 2*. Rule $C \in Th(M)$ is either a law and belongs to Law, or a sub-rule exists for it, which is true on M. This sub-rule shall be taken, then again it is either a law, or there is a sub-rule for it, true on M, and so on. As a result, the law is obtained, which is a sub-rule of C. Then, by virtue of theorem 1, it can be deduced from this law ∎

*Proof of the theorem 3*. The second exception follows from the definition. The first exception shall be considered. If rule $C \in L$ is of the form $C = (\Rightarrow A_0)$, it belongs to LP by definition. It shall be supposed that rule $C = (A_1 \& ... \& A_k \Rightarrow A_0)$ is a law within M. It shall be proved that $\eta(A_1 \& ... \& A_k) > 0$. If rule C is a law within M, the sub-rule $(A_2 \& ... \& A_k \Rightarrow \neg A_1)$ is not always true within M, and, thus, in some cases conjunction $A_2 \& ... \& A_k \& A_1$ is true, whence it follows that $\eta(A_2 \& ... \& A_k \& A_1) > 0$. Then conditional probabilities of all sub-rules are defined, since $\eta(A_{i_1} \& ... \& A_{i_h}) \geq \eta(A_1 \& ... \& A_k) > 0$ follows from $\{A_{i_1}, ..., A_{i_h}\} \subset \{A_1, ..., A_k\}$. It shall be proved that $\eta(C) = 1$.

$$\eta(C) = \eta(A_0 / A_1 \& ... \& A_k) = \eta(A_0 \& A_1 \& ... \& A_k) / \eta(A_1 \& ... \& A_k) =$$
$$\eta(A_0 \& A_1 \& ... \& A_k) / \eta(A_0 \& A_1 \& ... \& A_k) + \eta(\neg A_0 \& A_1 \& ... \& A_k).$$

Since rule C is true on M, there are no cases on M, when conjunction $(\neg A_0 \& A_1 \& ... \& A_k)$ is true and, hence, $\eta(\neg A_0 \& A_1 \& ... \& A_k) = 0$ and $\eta(C) = 1$.

It shall be proved that conditional probability of each sub-rule of rule $C = (A_1 \& ... \& A_k \Rightarrow A_0)$ is strictly less than $\eta(C) = 1$. Each sub-rule $(A_{i_1} \& ... \& A_{i_h} \Rightarrow L)$ of rule C is false on M, where L is either liter $\neg A$, or A, for the 1st and 2nd type sub-rules. It means that $\eta(A_{i_1} \& ... \& A_{i_h} \& \neg L) > 0$. It follows from the last inequality that:

$$\eta(L/A_{i_1} \& ... \& A_{i_h}) = \eta(A_{i_1} \& ... \& A_{i_h} \& L) / \eta(A_{i_1} \& ... \& A_{i_h}) =$$
$$\eta(A_{i_1} \& ... \& A_{i_h} \& L) / (\eta(A_{i_1} \& ... \& A_{i_h} \& \neg L) + \eta(A_{i_1} \& ... \& A_{i_h} \& L)) < 1.$$

Since $\eta(C) = 1$, rule C cannot be a sub-rule of any other probabilistic law, since in this case its conditional probability would be strictly less than conditional probability of this rule, what is impossible ∎

*Proof of the lemma 2.* If a probabilistic law has a form $C = (\Rightarrow A_0)$, it shall be verified, if it is the strongest probabilistic law. If it is true, then a semantic probabilistic inference is found, if no, there is a probabilistic law, for which this probabilistic law is a sub-rule. It shall be taken as the next rule of semantic probabilistic inference, and it shall be again verified, if it is the strongest law, and so on. A sub-rule, which is a probabilistic law, shall be found for probabilistic law $C = (A_1 \& ... \& A_k \Rightarrow A_0)$, $k \geq 1$. It always exists, since rule $C = (\Rightarrow A_0)$ is a probabilistic law. Since it is a sub-rule, its conditional probability will be less than a conditional probability of the rule as such. It shall be added as the previous rule of a semantic probabilistic inference, and the procedure shall be continued.

*Proof of the lemma 3.* Designations $a = \eta(G \& F \& H)$, $b = \eta(F \& H)$, $c = \eta(G \& F \& \neg H)$, $d = \eta(F \& \neg H)$ shall be introduced. Then an original inequality $\eta(G/F \& H) < \eta(G/F)$ shall be re-written as $a/b < (a+c)/(b+d)$, from which it follows that $(a+c)/(b+d) < c/d \Leftrightarrow \eta(G/F) < \eta(G/F \& \neg H)$ ∎

*Proof of the lemma 4.* Rule $C = (A_1 \& ... \& A_k \Rightarrow A_0)$ is either a probabilistic law, or there is a sub-rule $R' = (P_1 \& ... \& P_{k'} \Rightarrow A_0)$, $k' \geq 0$, $\{P_1,...,P_{k'}\} \subset \{A_1,...,A_k\}$, $k' < k$ such, that an inequality $\mu(R') \geq \mu(A)$ is satisfied. Similarly for rule $R'$, it is either a probabilistic law, or there is a sub-rule with identical properties for it ∎

*Proof of theorem 4.* It should be proved that for any sentence $H \in \Re(\Im)$, if $F(a) \& H(a), a \in A$ is true on M, inequality $\eta(G/F \& H) = \eta(G/F) = r$ is valid. From the condition of truth $F(a) \& H(a)$ on M, it follows that $\eta(F \& H) > 0$ and, hence, a conditional probability is defined.

A case shall be considered, when H is a liter (B or $\neg B$). The opposite shall be assumed that $\eta(G/F \& H) \neq r$. Then, as per lemma 3, one of inequalities $\eta(F \& B \Rightarrow G) > r$ or $\eta(F \& \neg B \Rightarrow G) > r$ shall be satisfied for one of the rules $(F \& B \Rightarrow G)$ or $(F \& \neg B \Rightarrow G)$. Then, as per lemma 4, a probabilistic law $C'$ exists, which is a sub-rule and has not lower conditional probability, then $\eta(C') > r$. Hence, by lemma 2, probabilistic law $C'$ belongs to some tree of semantic probabilistic inference and has a higher value of conditional probability than maximum specific rule $MS(G)$, predicting G, which contradicts to maximum specificity $MS(G)$.

A case shall be considered, when sentence H is a conjunction of two atoms $B_1 \& B_2$, for which a theorem has already been proved. Also, the opposite shall be assumed that one of inequalities $\eta(G/F\&B_1\&B_2)>r$, $\eta(G/F\&B_1\&\neg B_2)>r$, $\eta(G/F\&\neg B_1\&B_2)>r$, $\eta(G/F\&\neg B_1\&\neg B_2)>r$ is valid. Then, by lemma 4 and lemma 2, a probabilistic law $C'$ exists that belongs to the tree of a semantic probabilistic inference and is a sub-rule of one of these rules, and is such, as $\eta(C')>r$ is valid. However, it is impossible, since rule $C=(F\&H\Rightarrow G)$ is maximum specific. Hence, for all these inequalities only equality = or inequality < may take place. The last case is impossible due to the following equality

$$r = \frac{\eta(G\&F)}{\eta(F)} = \frac{\eta(G\&F\&B_1\&B_2) + \eta(G\&F\&\neg B_1\&B_2) + \eta(G\&F\&B_1\&\neg B_2) + \eta(G\&F\&\neg B_1\&\neg B_2)}{\eta(F\&B_1\&B_2) + \eta(F\&\neg B_1\&B_2) + \eta(F\&B_1\&\neg B_2) + \eta(F\&\neg B_1\&\neg B_2)}.$$

A case, when sentence H is a conjunction of several atoms or their negations, is called an induction.

In the general case, sentence $H\in\Re(\Im)$ may be presented as a disjunction of conjunctions of atoms or their negations. To complete the proof, suffice it to consider a case, when sentence H is a disjunction of two non-intersecting sentences $D\vee E$, $\eta(D\&E)=0$, for which a theorem is already proved, i.e.

$\eta(G\,/F\&D) = \eta(G\,/F\&E) = \eta(G\,/F) = r$.

Then $\eta(G\,/F\&(D\vee E)) = \dfrac{\eta(G\&F\&(D\vee E))}{\eta(F\&(D\vee E))} = \dfrac{\eta(G\&F\&D)+\eta(G\&F\&E)}{\eta(F\&D)+\eta(F\&E)} = r$ .

A disjunction case of multiple non-intersecting sentences is proved by induction ■

*Proof of proposition 2.* If L is compatible, atom $G\in L$ and its negation $\neg G\in L$ cannot exist simultaneously, since then $\eta(\&L)\leq \eta(G\&\neg G)=0$, where $\&L$ is a conjunction of liters from L ■

*Proof of the lemma 5.* A conditional probability shall be written as follows
$\eta(G/\overline{A}\&\neg\overline{B}) = \eta(G/\overline{A}\&(\neg B_1\vee...\vee\neg B_m))$.

Disjunction $\neg B_1\vee...\vee\neg B_m$ shall be presented as disjunction of conjunctions $\bigvee\limits_{i=(0,...,0)}^{i=(1,...,1,0)}(B_1^{i_1}\&...\&B_m^{i_m})$, where $i=(i_1,...,i_m)$, $i_1,...,i_m\in\{0,1\}$ zero means that there is a negation with the corresponding atom, and unity means that there is no negation. Disjunction does not involve set $(1,…,1)$, which corresponds to conjunction $B_1\&...\&B_m$.

Then a conditional probability $\eta(G/\overline{A}\&(\neg B_1\vee...\vee\neg B_m))$ shall be rewritten as

$\eta\left(G\,/\bigvee\limits_{i=(0,...,0)}^{i=(1,...,1,0)}(\overline{A}\&B_1^{i_1}\&...\&B_m^{i_m})\right)$.

It shall be proved that if $\eta(G/\overline{A}\&\neg\overline{B})>\eta(G/\overline{A})$, one of inequalities shall be satisfied

$\eta(G/\overline{A}\&B_1^{i_1}\&...\&B_m^{i_m})>\eta(G/\overline{A})$, $(i_1,...,i_m)\neq(1,...,1)$.

The opposite shall be assumed: all inequalities are simultaneously true

$\eta(G/\overline{A}\&B_1^{i_1}\&...\&B_m^{i_m})\leq\eta(G/\overline{A})$, $(i_1,...,i_m)\neq(1,...,1)$

in those cases, when $\eta(\overline{A}\&B_1^{i_1}\&...\&B_m^{i_m})>0$. Since $\eta(\overline{A}\&\neg\overline{B})>0$, there are cases, when

$\eta(\overline{A}\&B_1^{i_1}\&...\&B_m^{i_m})>0$.

Then

$$\eta(G \& \bar{A} \& B_1^{i_1} \& ... \& B_m^{i_m}) \leq \eta(G/\bar{A})\eta(\bar{A} \& B_1^{i_1} \& ... \& B_m^{i_m}), \quad (i_1,...,i_m) \neq (1,...,1),$$

$$\eta\left(G / \bigvee_{i=(0,...,0)}^{i=(1,...,1,0)} (\bar{A} \& B_1^{i_1} \& ... \& B_m^{i_m})\right) = \frac{\eta\left(\bigvee_{i=(0,...,0)}^{i=(1,...,1,0)} (G \& \bar{A} \& B_1^{i_1} \& ... \& B_m^{i_m})\right)}{\eta\left(\bigvee_{i=(0,...,0)}^{i=(1,...,1,0)} (\bar{A} \& B_1^{i_1} \& ... \& B_m^{i_m})\right)} =$$

$$\frac{\sum_{i=(0,...,0)}^{i=(1,...,1,0)} \eta(G \& \bar{A} \& B_1^{i_1} \& ... \& B_m^{i_m})}{\sum_{i=(0,...,0)}^{i=(1,...,1,0)} \eta(\bar{A} \& B_1^{i_1} \& ... \& B_m^{i_m})} \leq \frac{\eta(G/\bar{A}) \sum_{i=(0,...,0)}^{i=(1,...,1,0)} \eta(\bar{A} \& B_1^{i_1} \& ... \& B_m^{i_m})}{\sum_{i=(0,...,0)}^{i=(1,...,1,0)} \eta(\bar{A} \& B_1^{i_1} \& ... \& B_m^{i_m})} = \eta(G/\bar{A}),$$

that contradicts to inequality $\eta(G/\bar{A}\&\neg B) > \eta(G/\bar{A})$. Therefore, this assumption is not valid, and there is a rule as follows

$$\bar{A} \& B_1^{i_1} \& ... \& B_m^{i_m} \Rightarrow G, \quad (i_1,...,i_m) \neq (1,...,1),$$

that has a strictly higher assessment of conditional probability than A ∎

***Proof of theorem 5***: It should be proved that, every time a rule from $P \subset MSR$ is applied, a compatible set of liters is once again obtained. The opposite shall be assumed that, when applying some rule $A = (A_1 \& ... \& A_k \Rightarrow G), \{A_1,...,A_k\} \subset L, k > 1$ to a set of liters $L = \{L_1,...,L_k\}$, a liter G shall be obtained, for which $\mu(L_1 \& ... \& L_n \& G) = 0$.

Since for rules inequalities $\eta(G/A_1 \& ... \& A_k) > \eta(G)$, $\eta(A_1 \& ... \& A_k) > 0, \eta(G) > 0$ are satisfied, then $\eta(G \& A_1 \& ... \& A_k) > \eta(G)\eta(A_1 \& ... \& A_k) > 0$.

Negations of liters $\{B_1,...,B_t\} = \{L_1 \& ... \& L_n\} \setminus \{A_1,...,A_k\}$ shall be added to rule A, and rule $(A_1 \& ... \& A_k \& \neg(B_1 \& ... \& B_t) \Rightarrow G)$ shall be obtained. The following designations shall be made

$\&A_i = A_1 \& ... \& A_k$, $\&B_j = B_1 \& ... \& B_t$, $\&L = L_1 \& ... \& L_n$.

As it was assumed, $\mu(L_1 \& ... \& L_n \& G) = 0$ and $\eta(\&A_i \& (\&B_j)) = \eta(\&L) > 0$. It shall be proved that in this case $\eta(\&A_i \& \neg(\&B_j)) > 0$. The opposite shall be assumed that $\eta(\&A_i \& \neg(\&B_j)) = 0$, then

$\eta(G \& (\&A_i) \& \neg(\&B_j)) \leq \eta(\&A_i \& \neg(\&B_j)) = 0.$

Whence it follows that

$0 = \mu(\&L \& G) = \eta(G \& (\&A_i) \& (\&B_j)) = \eta(G \& (\&A_i) - \eta(G \& (\&A_i) \& \neg(\&B_j))) =$
$\eta(G \& A_1 \& ... \& A_k) > \eta(G)\eta(A_1 \& ... \& A_k) > 0.$

A comtradiction is obtained. Then

$$\eta(G/\&A_i \& \neg(\&B_j)) = \frac{\eta(G \& (\&A_i) \& \neg(\&B_j))}{\eta(\&A_i \& \neg(\&B_j))} = \frac{\eta(G \& (\&A_i)) - \eta(G \& (\&A_i) \& (\&B_j))}{\eta(\&A_i) - \eta(\&A_i \& (\&B_j))} =$$

$$\frac{\eta(G \& (\&A_i)) - \eta(\&L \& G)}{\eta(\&A_i) - \eta(\&A_i \& (\&B_j))} = \frac{\eta(G \& (\&A_i))}{\eta(\&A_i) - \eta(\&A_i \& (\&B_j))} > \frac{\eta(G \& (\&A_i))}{\eta(\&A_i)} = \eta(G/A_1 \& ... \& A_k).$$

Then, by virtue of lemmas 2, 4, 5, it shall be found that there exists a probabilistic law with a higher conditional probability than rule A, which contradicts to a maximum specificity of rule A ∎